\documentclass{article}
\usepackage{kr}

\usepackage{times}
\usepackage{soul}
\usepackage{url}
\usepackage[hidelinks]{hyperref}
\usepackage[utf8]{inputenc}
\usepackage[small]{caption}
\usepackage{graphicx}
\usepackage{amsmath}
\usepackage{amsthm}
\usepackage{booktabs}
\usepackage{algorithm}
\usepackage{algorithmic}
\usepackage{comment}
\usepackage{xspace}
\usepackage{booktabs}
\usepackage{fancyvrb}
\usepackage{listings}
\usepackage{fvextra}
\usepackage{multirow}
\usepackage{array}
\usepackage{tablefootnote}
\usepackage{subfigure}
\usepackage{subcaption}

\newcommand{\spara}[1]{\smallskip\noindent{\bf #1}}

\makeatletter
\newcommand\footnoteref[1]{\protected@xdef\@thefnmark{\ref{#1}}\@footnotemark}
\makeatother

\newcommand{\llm}{\ensuremath{\mathcal{L}}\xspace}
\newcommand{\dataset}{\ensuremath{\mathcal{D}}}
\newcommand{\template}{\ensuremath{\mathcal{T}}}
\newcommand{\prompts}{\ensuremath{\mathcal{P}}}

\newcommand{\answers}{\ensuremath{\mathcal{A}}}

\newcommand{\llmprob}[2]{\ensuremath{P_\llm\left(#1|#2\right)}}
\newcommand{\probclass}{\ensuremath{\mathcal{C}}}

\newcommand{\x}{\ensuremath{x}\xspace}
\newcommand{\y}{\ensuremath{y}\xspace}

\newcommand{\yreal}{\ensuremath{y^*}\xspace}

\newcommand{\clingo}[1]{\ensuremath{s(#1)}\xspace}

\newcommand{\as}[1]{\ensuremath{\mathit{AS}(#1)}\xspace}

\usepackage{xcolor}
\usepackage{xargs}
\usepackage[colorinlistoftodos,prependcaption,textsize=small]{todonotes}
\usepackage[framemethod=tikz]{mdframed}

\newcommandx{\unsure}[2][1=]{\todo[linecolor=red,backgroundcolor=red!25,bordercolor=red,caption={},#1]{#2}}
\newcommandx{\change}[2][1=]{\todo[linecolor=blue,backgroundcolor=blue!25,bordercolor=blue,caption={},#1]{#2}}
\newcommandx{\info}[2][1=]{\todo[linecolor=green,backgroundcolor=green!25,bordercolor=green,caption={},#1]{#2}}
\newcommandx{\improvement}[2][1=]{\todo[linecolor=purple,backgroundcolor=purple!25,bordercolor=purple,caption={},#1]{#2}}
\newcommandx{\discussion}[2][1=]{\todo[linecolor=blue,backgroundcolor=yellow!25,bordercolor=purple,caption={},#1]{#2}}

\newcommandx{\thiswillnotshow}[2][1=]{\todo[disable,#1]{#2}}

\newcommand{\mytilde}{$\sim$}
\newcommand{\derives}{\verb~:-~}

\newmdenv[innerlinewidth=0.5pt, roundcorner=4pt,linecolor=gray,backgroundcolor=gray!25,innerleftmargin=6pt,innerrightmargin=6pt,innertopmargin=6pt,innerbottommargin=6pt]{problem}

\newtheorem*{example}{Example}

\title{LLASP: Fine-tuning Large Language Models for Answer Set Programming}

\author{%
Erica Coppolillo$^{1,2}$\and
Francesco Calimeri$^{1,3}$\and
Giuseppe Manco$^2$\and
Simona Perri$^1$ \and
Francesco Ricca$^1$ \\
\affiliations
$^1$University of Calabria, Via Bucci 28c, Rende - Italy\\
$^2$ICAR-CNR, Via Bucci 8/9C, Rende - Italy\\
$^3$DLVSystem Srl, Viale della Resistenza 19C, Rende - Italy \\
\emails
\{erica.coppolillo, francesco.calimeri, simona.perri, francesco.ricca\}@unical.it, 
giuseppe.manco@icar.cnr.it
}

\begin{document}

\maketitle


\begin{abstract}
  Recently, Large Language Models (LLMs) have showcased their potential in various natural language processing tasks, including code generation. However, while significant progress has been made in adapting LLMs to generate code for several imperative programming languages and tasks, there remains a notable gap in their application to declarative formalisms, such as Answer Set Programming (ASP). In this paper, we move a step towards exploring the capabilities of LLMs for ASP code generation. First, we perform a systematic evaluation of several state-of-the-art LLMs. Despite their power in terms of number of parameters, training data and computational resources, empirical results demonstrate inadequate performances in generating correct ASP programs. Therefore, we propose LLASP, a fine-tuned lightweight model specifically trained to encode fundamental ASP program patterns. To this aim, we create an ad-hoc dataset covering a wide variety of fundamental problem specifications that can be encoded in ASP. Our experiments demonstrate that the quality of ASP programs generated by LLASP is remarkable. This holds true not only when compared to the non-fine-tuned counterpart but also when compared to the majority of eager LLM candidates, particularly from a semantic perspective.
All the code and data used to perform the experiments are publicly available: \url{https://anonymous.4open.science/r/LLASP-D86C/}.
\end{abstract}



\section{Introduction}

Answer Set Programming (ASP) ~\cite{DBLP:journals/cacm/BrewkaET11,Lifschitz19} is a pivotal tool for knowledge representation and reasoning, offering a range of benefits that contribute to its significance in various domains. One notable advantage is ASP capacity to handle knowledge in a concise and intuitive manner, allowing users to express complex problems with relative ease~\cite{DBLP:books/daglib/0040913}. 
ASP provides a declarative approach for expressing knowledge and solving combinatorial optimization, planning, and reasoning tasks~\cite{DBLP:series/synthesis/2012Gebser}. 
Its non-monotonic nature allows for the representation of incomplete information and default reasoning, essential for capturing real-world scenarios. Moreover, ASP expressiveness enables the integration of various types of knowledge, including rules, constraints, and preferences, facilitating flexible problem-solving~\cite{Lifschitz19}. 

This formalism is typically regarded as a specialized tool utilized by domain experts and knowledge engineers familiar with specific problem domains. These individuals leverage ASP as a declarative language to articulate problem constraints, rules, and preferences succinctly and intuitively. However, despite the language simplicity and conciseness, coding in ASP can be an overwhelming obstacle for non-expert users, due to lack of familiarity with declarative paradigms and their intrinsic semantics, as well as potential complexity of application domains. 
Under this perspective, it is important to develop tools that enhance the efficiency and automation of creating ASP programs. These tools would minimize the disparity between natural language specifications and the corresponding ASP source code, facilitating smoother development processes.~\cite{DBLP:conf/bionlp/ErdemY09,DBLP:conf/inap/FangT17,DBLP:journals/tplp/Schwitter18,DBLP:journals/tplp/CarusoDMMR24}. 

In this respect, the recent advancements in Artificial Intelligence and Machine/Deep Learning have led to the emergence of Large Language Models (LLMs) as indispensable assets across various natural language applications and tasks~\cite{survey,llms-survey}. Notably, in the domain of automatic text-to-code translation, LLMs are employed to generate programs from natural language inputs. These models are trained on extensive datasets encompassing code repositories, technical forums, coding platforms, documentation, and relevant web data associated with programming. 
Such comprehensive training equips LLMs with the ability to grasp the nuances of code context, 
thereby enhancing the accuracy of contextually relevant code generation.

Indeed, a natural prosecution in this context is the investigation of LLMs-based approaches aimed at automating code generation within the context of declarative programming. This research challenge aims at narrowing the gap between natural language specifications and ASP source code, and aligns with the broader trend of leveraging language models for code automation. Further, leveraging LLMs for automating ASP program generation offers several advantages, which include: better semantic specification through high-level natural language descriptions of their problem domains; contextual awareness and interoperability, which simplifies the process of specifying ASP programs, making it accessible to a wider range of users without deep programming expertise; and enhanced efficiency and productivity. Surprisingly, the current literature exhibits a significant discrepancy in this respect, even if some initial attempts have been made~\cite{DBLP:conf/kr/IshayY023,ijcai24-borroto}.

In an effort to rectify this gap, the present work proposes a systematic study on the capabilities of currently available LLMs to automate the generation of ASP programs. In our study, we identify some structural properties (classes) of ASP programs specifically tailored to fundamental tasks, and correlate them with the generation capabilities of the currently available LLM architectures. Inspired by such a characterization, we further devise a methodology for obtaining more accurate ASP encodings. Our approach consists in the construction of a custom training dataset, specifically tailored for the identified relevant tasks, and in fine-tuning an LLM instance that can be successfully exploited for ASP coding. Through an extensive experimentation, we show that even a fine-tuned lightweight base-model is more effective than heavy-sized LLMs. As a result, the proposed methodology offers a pathway towards fine-tuned models ready for practical deployment.



Our contributions can be summarized as follows. \begin{itemize}
\item 
    To the best of our knowledge, we are the first to extensively compare state-of-the-art LLMs in terms of ASP code generation. Our findings suggest that, despite their potential, LLMs still underperform in terms of syntactic and semantic program correctness.
\item 
    We show that a tailored training strategy, even if applied on a lightweight LLM, can outperform state-of-the art large-size models. To address this, we curated a comprehensive dataset and used it to produce LLASP, the fine-tuned version of a Gemma 2B base-model specifically trained to capture ASP fundamental patterns. 
\item 
    Via a comprehensive experimental evaluation, we prove that LLASP generate ASP programs that are highly valuable in terms of both syntactic and semantic accuracy, overcoming significantly larger and more powerful LLMs especially under a semantic perspective. 
\item 
    Finally, we provide a discussion on the current limitations and potential directions for extensions of our proposed approach, inspired by the results of additional experiments.
\end{itemize}

The rest of the paper is structured as follows. Section~\ref{sec:related} discusses the state of the art in automated code generation and the drawbacks of current approaches focused on ASP. In Section~\ref{sec:preliminaries}, we introduce the basic concepts, and in Section~\ref{sec:design} we propose our methodology and design patterns for the evaluation and fine-tuning of the model. Section~\ref{sec:exp} reports the experiments and discusses our findings. Finally, in Section~\ref{sec:conc} we set some pointers for future research.

\section{Related Work}\label{sec:related}
The advantages of automating code synthesis are well-recognized in the literature~\cite{DBLP:journals/software/ErnstBM22,kalliamvakou2022research,DBLP:journals/corr/abs-2302-06590,DBLP:journals/jss/DakhelMNKDJ23}, 
and nearly all mainstream programming languages are nowadays supported by automated program composition tools~\cite{DBLP:journals/corr/abs-2107-03374}.
In this area, Large Language Models (LLMs) play a central role, and their performance has been extensively compared in the literature~\cite{evaluation-code,codet5}. 
The positive impact of fine-tuning models for code generation of imperative programming is also established~\cite{llamoco}. 

In the context of declarative programming, a widely recognized research objective is to create tools that streamline and automate the development of Answer Set Programming (ASP) programs. The 
goal is to bridge the gap between natural language specifications and ASP source code~\cite{DBLP:conf/bionlp/ErdemY09,DBLP:conf/inap/FangT17,DBLP:journals/tplp/Schwitter18,DBLP:journals/tplp/CarusoDMMR24}. 
Initial proposals focused on automating the resolution of logic puzzles presented in simplified English by translating their descriptions into ASP~\cite{DBLP:conf/aaaifs/BaralD11}, employing $\lambda$-calculus and probabilistic combinatorial categorical grammars.
Later, several efforts have been spent in the development of Controlled Natural Languages (CNLs)~\cite{DBLP:journals/coling/Kuhn14} for ASP programs. 
CNLs represent subsets of full natural languages, featuring restricted grammar and vocabulary. 
Among them, the BIOQUERYCNL~\cite{DBLP:conf/bionlp/ErdemY09} defines the grammatical structure of a CNL and algorithm for transpling queries into ASP; 
Fang et al. introduced a CNL approach for ASP that leverages LANA annotations, that was implemented in the SeaLion IDE~\cite{DBLP:conf/inap/FangT17}. 
In 2018, Schwitter developed a CNL called PENG\textsuperscript{ASP} for specifying and verbalizing answer set programs~\cite{DBLP:journals/tplp/Schwitter18}. 
More recently, Dodaro et al. introduced CNL2ASP, an extensive publicly-available tool for converting controlled natural language into ASP programs~\cite{DBLP:journals/tplp/CarusoDMMR24}. 
On the one hand, the CNLs provide the programmer with a language that is more similar to the natural one, thus greatly reducing the barrier of coding in ASP code;
on the other hand, CNLs still have their own syntactic constraints that do not free completely the user from structuring the sentence according to a formal syntax. 

Even powerful language tools such as LLMs have been exploited in conjunction with the ASP formalism in several ways. \cite{DBLP:conf/nips/NyeTTL21} introduced a dual-system model based on GPT-3, comprising neural System 1 and logical System 2, which generates semantic parsers from natural language sentences and integrates them with reasoning modules. In a similar vein, \cite{DBLP:conf/acl/YangI023} proposed that LLMs like GPT-3 can function as few-shot semantic parsers, transforming natural language into logical forms for ASP without necessitating distinct retraining for diverse question-answering tasks. \cite{DBLP:conf/kr/IshayY023} utilized LLMs with prompt engineering to obtain ASP solutions for logic puzzles, leveraging the logic puzzle dataset from~\cite{DBLP:conf/aaai/MitraB16}. 
Further, the effective utilization of Large Language Models in combination with ASP to perform a number of natural language understanding tasks has been explored and implemented in the STAR framework~\cite{reliable-nlu}. 

Despite the growing interest in this area, however, a notable gap still persists in the literature regarding the adoption of LLMs for code synthesis for declarative programming languages, such as ASP. In 2024, Borroto et al. moved the first steps in this regard. 
Their method, implemented in the NL2ASP tool, constructs ASP programs from natural language specifications through a two-step architecture~\cite{ijcai24-borroto}. 
Indeed, NL2ASP first applies neural machine translation~\cite{DBLP:journals/jair/Stahlberg20} to transform natural language into the statements of the CNL by Dodaro et al.; then, in the second step, the produced sentences are used to generate ASP code by means of the CNL2ASP tool. 
NL2ASP has been implemented with two Transformer-based models for NMT tasks, i.e., T5-small, and Bart-base, and demonstrated promising performance. 

Notably, there are some key differences between the approach proposed in this paper and the cited proposal.
First, NL2ASP currently focuses on generating ASP programs for solving graph-related problem specifications, whereas the approach presented in this paper does not focus on a specific domain. Rather, it targets the generation of ASP programs suitable for solving general problems, starting from some fundamental specific ASP patterns, as we will show later. 
Another important difference is that our approach does not rely on an intermediate format: rather, it aims at producing ASP code directly from natural language specifications. 
Finally, it is worth observing that, even though 
low performance of LLMs in generating ASP code was already preliminarily observed~\cite{ijcai24-borroto}, a systematic and exhaustive empirical analysis assessing this behaviour was still missing, and is provided in this paper.

\section{Preliminaries}\label{sec:preliminaries}

\subsubsection{Large Language Models.} A Large Language Model \llm can be formally described as a  function $f$, which stochastically maps input sequences of tokens $\x = [x_1, x_2, ..., x_n]$ to an output sequence $\y = [y_1, y_2, ..., y_m]$, where $n$ represents the length of the input sequence and $m$ represents the length of the output sequence. The model defines \llmprob{\y}{\x}, i.e., the probability distribution of $\y$ given $\x$, capturing complex patterns and relationships in natural language.
Since $\x \in V^*$ and $\y \in W^*$, where $V$, $W$ represent vocabularies of tokens, both $\x$ and $\y$ should be compliant with respect to some grammar. The $f$ function is finally defined based on a sampling of $\y$ from $\llmprob{\cdot}{\x}$.

The above probability is computed by relying on the Transformer Encoder-Decoder architecture introduced in~\cite{attention-is-all}. In more details, the input sequence of tokens is first converted into dense vector representations, which encode context similarity into geometrical closeness and also embed positional information. These embeddings then feed the Transformer layers.  
The Encoder consists of multiple identical layers, each with self-attention to weigh word importance, and a fully connected feedforward network for independent position transformations. Similarly, Decoder layers feature self-attention with masking, Encoder-Decoder Attention for input focus during token generation, and position-wise feedforward networks.
Finally, the output of the Transformer Decoder is passed through a linear layer followed by a softmax activation function, which generates a probability distribution over the vocabulary of possible output tokens. 

The latter implements the function $\llmprob{\y_i}{\x, \y_{:i-1}}$ upon which the sampling process can be devised as a sequence of steps. Here, $\y_{:i-1}$ represents the prefix of $\y$ up to position $(i-1)$. 
Several architectural variants have been proposed (e.g., Causal-Decoder, Prefix Decoder, Autoregressive, Mixture Variants~\cite{comprehensive-overview,survey}). The underlying architecture of the LLM is not deeply investigated in this paper, since it is orthogonal to our study.


The training of $\llm$ is usually accomplished in steps. The first step consists in a pre-training task which is based on next-word prediction. In practice, given $\x$ and a partial response $\y_{:i-1}$, the task is to learn to correctly predict $\y_i$. The next phase is called supervised fine-tuning (SFT), and it consists in refining \llm to predict the whole $\y$ given $\x$, on a large corpora $\mathcal{D} =\left\{(\x^{(1)}, \y^{(1)}), \ldots, (\x^{(N)}, \y^{(N)})\right\}$, where $N$ is the number of training pairs. The refinement is based on the loss $\ell\left(\x^{(i)},\llmprob{\y^{(i)}}{\x^{(i)}}\right)$, usually defined in terms of cross-entropy: 
$$
\ell\left(\y^{(i)},\llmprob{\y^{(i)}}{\x^{(i)}}\right) = \sum_j y^{(i)}_j \log \llmprob{y^{(i)}_j}{\x^{(i)},\y^{(i)}_{:j-1}}
$$
In the following, we shall refer to this formulation of the loss, even though alternative formulations are possible.


\subsubsection{Answer Set Programming.} 
ASP is a powerful declarative formalism for Knowledge Representation and Reasoning
that gained increasing interest for its high expressive power, and the availability of solid and effective implementations~\cite{DBLP:conf/ijcai/GebserLMPRS18}. It is fully declarative (i.e., the ordering of literals and rules is immaterial), and the encoding of a large variety of problems is simple and concise. %

In the following, we briefly recall the syntax of the language, focusing on the aspects relevant to this work, and provide an intuitive semantics, revisiting the concept of \textit{answer sets} that will play a crucial role in the validation process outlined in Section~\ref{sec:validazione-esperimenti}.
%
For further details and complete references to advanced ASP features, the reader may refer to the vast literature~\cite{DBLP:journals/cacm/BrewkaET11,DBLP:conf/rweb/EiterIK09,DBLP:journals/ai/CalimeriGMR16}.

A \textit{term} is either a constant or a variable. 
An \textit{atom} is an expression $p(t_1, \dots,t_n)$, where \textit{p} is a predicate of arity \textit{n} and $t_1, \dots,t_n$ are terms. 
A program $P$ is a finite set of rules, constructs like $\textit{Head}\ \derives\ \textit{Body}$ where $\textit{Head}$ is a disjunction of atoms and $\textit{Body}$ is a conjunction of literals; 
more formally, a rule is of the form
$$\alpha_1 | \cdots | \alpha_k\ :-\ \beta_1,\dots,\beta_n, \ \mathit{not} \  \beta_{n+1},\dots,\ \mathit{not} \ \beta_m.$$ 
where $m \geq 0, k \geq 0$; 
$\alpha_1, \cdots, \alpha_k$ and $\beta_1,\dots,\beta_m$ are atoms. 
A rule is interpreted according to common sense principles: roughly, its intuitive semantics corresponds to an implication. 
When a predicate $p$ occurs in the head of a rule $r$, we say that $r$ defines $p$. 
Rules featuring an atomic formula in the head and an empty body are used to represent information known to be certainly true and indeed called {\em facts}. 
A rule with empty head is called \textit{(strong) constraint}, and intuitively expresses conditions that must be satisfied; basically, it is forbidden that all the literals in the body of a constraint are true.
ASP also supports {\em weak constraints}, i.e, special rules with empty heads which allow to express preferences, and hence to deal with optimization problems. 
A weak constraint of the form 
$$:\sim l_1,\cdots,l_n.\ \ \ [w@l,t_1,...,t_m]$$  
associates literals $l_1,\cdots,l_n$ with a weight $w$, a level $l$, and additional terms $t_1,\cdots,t_m$ for $m \geq 0$. 
Intuitively, weak constraints allow to express conditions that should be satisfied, but not necessarily have to be; 
if a weak constraint is violated (i.e., its body is true), then the weight $w$ (``cost'') is paid at priority level $l$. 
%
Finally, a program (a rule, an atom) that does not contain variables is said to be \textit{ground}.  

Models are defined over ground programs:
a {\em model} for a program $P$ is a subset of all possible ground atoms that satisfies all rules in $P$.
The semantics of ASP programs is given in terms of special models called {\em Answer Sets}: according to this semantics, an ASP program may have several alternative answer sets (but possibly none), each corresponding to a possible view of the world~\cite{DBLP:journals/ngc/GelfondL91,DBLP:journals/cacm/BrewkaET11}. 
In ASP, a computational problem is typically solved by modeling it via a program consisting of a collection of rules along with a set of facts representing the instance at hand; then, solutions are found by computing the intended answer sets, according to the so-called answer set semantics. Answer sets correspond one-to-one to the solutions of the given instance of the modeled problem: if a program has no answer sets, the corresponding instance has no solutions.
If weak constraints are present, optimal answer sets are those minimizing the sum of weights of the violated weak constraints in the highest priority level. Among these, the optimal sets are those that also minimize the sum of weights of the violated weak constraints at the next lower level, and so forth. 
\color{black}

\section{Methodology and Knowledge Design}
\label{sec:design}

In the present work, we address a twofold objective. First, we focus on pre-trained general-purpose Large Language Models, not specifically tailored for ASP generation. Thus, given an LLM \llm and a natural language specification \x of a problem, we aim at evaluating the robustness of \llm in sampling  $\y \sim  \llmprob{\cdot}{x}$, such that \y is an ASP encoding compliant to \x. 
Our second objective revolves around exploring how fine-tuning impacts the quality of generation, particularly when dealing with small-sized models.  
For a comprehensive evaluation, we need to 
consider various aspects:
\begin{itemize}
    \item \textbf{Dataset diversity and complexity.} We need to assemble a dataset encompassing a wide range of problem descriptions (\x) covering different domains, complexities, and lengths. This diversity ensures that the evaluation captures \llm's performance across various contexts and complexity levels, thus helping in discerning \llm's efficacy in handling various ASP encoding challenges.
    
    \item \textbf{ASP compliance and robustness.} Since each problem description \x will be paired with a corresponding ASP encoding $\y ~\sim P_\llm(\cdot|x)$, we need to verify whether \y adhere to the ASP syntax, as well as its fidelity to the original problem description \x. Metrics such as syntax accuracy, semantic coherence, and adherence to problem constraints should be considered.

    \item \textbf{Impact of model size.} 
    We need to investigate how the size and architecture of the \llm influence its ability to generate accurate ASP encodings. 
    This analysis should involve comparing the performance of different \llm variants (in particular, smaller vs. larger models) and assessing any trade-offs between model size and encoding quality.

    \item \textbf{Generalizability assessment.} We need to examine whether \llm, trained on a diverse set of tasks, can generalize well to new problem domains not encountered during training. This  helps devise the extent to which \llm's capabilities extend beyond its training data and whether it can effectively tackle novel ASP encoding tasks.
\end{itemize}

We concentrate on the first aspect for now, addressing the others in the subsequent section.
Here, we need to define an ad-hoc approach for tuning \llm towards the production of ASP programs.
In this regard, we take inspiration from the typical learning process experienced by humans while trained on the task of encoding problems into ASP programs. 
In particular, the process is based on modeling simple patterns encoding basic pieces of knowledge.
For instance, how to represent with ASP a Cartesian Product, the Join between the extensions of two predicates, simple guesses, conditions on specific values, transitive closures, and so on.
In principle, more complex patterns can be devised in a compositional way, starting from these basic building blocks to produce larger sets of rules that, in turn encode more complex knowledge. This approach to ASP generation is in general effective: the declarative nature of the formalism makes the order between rules in a program, and the order of literals in the bodies, immaterial; thus, in many cases, a program representing the desired solution can be composed by just adding subprograms representing additional knowledge. 

According to this idea, we defined a template set $\template = (\prompts, \answers)$, 
where $\prompts$ is a set of descriptions of basic tasks expressed in natural language and $\answers$ is a set of text snippets consisting of corresponding (\textit{gold}) ASP programs that encode them. 
%
%
In practice,  $\template$ is partitioned into  $\{\template_{\probclass^1}, \template_{\probclass^2}, ..., \template_{\probclass^k}\}$, where $\probclass^i$, represents the $i$-th task type and $k$ is the total number of modeled task types.
For instance, if $\probclass^1$ refers to a transitive closure task, $\template_{\probclass^1}$ consists of pairs ($\prompts_{\probclass^1},\answers_{\probclass^1}$) featuring a description in natural language and one among the corresponding suitable gold ASP encodings, respectively. 
Each template (and its gold ASP counterpart) exhibits some placeholders to be instantiated. Such placeholders represent predicates, labels and values within the ASP encoding.  

Below, we provide a brief overview of the fundamental tasks we examined. For each task, we offer an informal description, the template utilized, and an example of instantiated prompt along with its corresponding ASP encoding.
%


\spara{Guessing Assignments.}
We started with considering a typical task where all elements of a given set must be assigned to a unique label picked from a fixed set. 
A common way for representing this in ASP is to use disjunction in order to express a so-called ``guess''. 
A possible pair associating a prompt with a suitable ASP encoding is reported next.
%
\begin{scriptsize}
\begin{Verbatim}[frame=single, breaklines=true, breakanywhere=true, breaksymbol=$\quad$]
Template: Write an ASP program for the following problem. Assign exactly a label among a given set of labels to a set of elements. The set of elements is expressed by predicate [PREDICATE]. The labels are [LABEL]+.

Prompt: Write an ASP program for the following problem. Assign exactly a label among a given set of labels to a set of elements. The set of elements is expressed by predicate city. The labels are moscow,rome,dubai.

Encoding: assign(X,"moscow") | assign(X,"rome") | assign(X,"dubai") :- city(X).
\end{Verbatim}
\end{scriptsize}

\spara{Expressing Constraints.}
Another typical task is to express conditions that must be fulfilled.
Within ASP, this is commonly expressed by mean of (classical/strong) constraints, possibly associated with auxiliary rules if needed. 
A possible pair associating a prompt asking for preventing a specific assignment with a suitable ASP encoding is reported next.
\begin{scriptsize}
\begin{Verbatim}[frame=single, breaklines=true, breakanywhere=true, breaksymbol=$\quad$]
Template: Write an ASP program for the following problem. Prevent the predicate [PREDICATE] with value [VALUE] from having label [LABEL].

Prompt: Write an ASP program for the following problem. Prevent the predicate car with value 11 from having label "red".

Encoding:  :- assign(11,"red").
\end{Verbatim}
\end{scriptsize}

\spara{Generating Combinations.}
The need for generating all combinations of elements from two different sets (i.e., defining the Cartesian product) can be in general fulfilled in ASP by means of a simple rule. 
A possible pair associating a prompt with a suitable ASP encoding is reported next.
\begin{scriptsize}
\begin{Verbatim}[frame=single, breaklines=true, breakanywhere=true, breaksymbol=$\quad$]
Template: Write an ASP program for the following problem. Generate all the combinations of elements from two sets. The two sets are represented by predicates [PREDICATE_1] and [PREDICATE_2].

Prompt: Write an ASP program for the following problem. Generate all the combinations of elements from two sets. The two sets are represented by predicates city and airport.

Encoding: combination(X,Y) :- city(X), airport(Y).
\end{Verbatim}
\end{scriptsize}

\spara{Joins.}
Joins over the elements of two different sets according to specific matching criteria over features can be typically defined in ASP via rules featuring proper body literals having a variable in common.
A possible pair associating a prompt with a suitable ASP encoding is the following.
\begin{scriptsize}
\begin{Verbatim}[frame=single, breaklines=true, breakanywhere=true, breaksymbol=$\quad$]
Template: Write an ASP program for the following problem. Consider predicate [PREDICATE_1] having fields [LABEL]+ and the predicate [PREDICATE_2] having fields [LABEL]+. Define a predicate [PREDICATE_1]_[PREDICATE_2] that associates to each [PREDICATE_1] the [LABEL] of [PREDICATE_2].

Prompt: Write an ASP program for the following problem. Consider predicate "owner" having fields "ID","surname", "name","restaurantID", and the predicate "restaurant" having fields "ID","description". Define a predicate "owner_restaurant" that associates to each owner the description of restaurant.

Encoding: owner_restaurant(X,Z) :- owner(X,_,_,Y), 
                                   restaurant(Y,Z).
\end{Verbatim}
\end{scriptsize}

\spara{Transitive Closure.}
Transitive closure is a fundamental tool for defining the structure of relationships in a variety of contexts as it catches not only direct relationships among elements, but also those that are indirectly resulting from a chain of relations.
Expressing this with ASP in general requires to use more than a single rule. 
A possible pair associating a prompt describing a transitive closure with a suitable ASP encoding is reported next; in this case, the encoding make use of two rules: one for defining the direct relations, and another (relying on recursion) for catching the indirect ones.
\begin{scriptsize}
\begin{Verbatim}[frame=single, breaklines=true, breakanywhere=true, breaksymbol=$\quad$]
Template:  Write an ASP program for the following problem. Define predicate [PREDICATE_1] as the transitive closure of predicate [PREDICATE_2].

Prompt: Write an ASP program for the following problem. Define predicate "arrivals" as the transitive closure of predicate "flight".

Encoding: 
    arrivals(X,Y) :- flight(X,Y).
    arrivals(X,Y) :- flight(X,Z),arrivals(Z,Y).
\end{Verbatim}
\end{scriptsize}

\spara{Expressing Preferences.}
In order to use ASP for solving optimization problems, one must be able to express, to several extents, preferences over the set of admissible solutions;
this is usually done by encoding programs featuring {\em weak constraints}. 
A possible pair associating a prompt asking for expressing a specific preference with a suitable ASP encoding is reported next.
\begin{scriptsize}
\begin{Verbatim}[frame=single, breaklines=true, breakanywhere=true, breaksymbol=$\quad$, commandchars=\\\{\}]
Template: Write an ASP program for the following problem. I would prefer that predicate [PREDICATE] with value 18 is not associated with [LABEL]. If this occurs, it costs [COST] at level [LEVEL].

Prompt: Write an ASP program for the following problem. I would prefer that predicate house with value 18 is not associated with "flat". If this occurs, it costs 2 at level 2.

Encoding: :\mytilde assign(18,"flat"). [2@2]
\end{Verbatim}
\end{scriptsize}

\noindent \textit{Filtering.}
When encoding ASP programs, it is frequently necessary to apply filters to the extensions of certain predicates based on specific requirements.
We report next some common types of filtering criteria.

\spara{Filtering by values.}
The first type of filtering criteria consists in selecting the portion of the extension of a predicate that match with a specific value. 
A possible pair associating a prompt with a suitable ASP encoding is reported next.
\begin{scriptsize}
\begin{Verbatim}[frame=single, breaklines=true, breakanywhere=true, breaksymbol=$\quad$]
Template: Write an ASP program for the following problem. Select all values associated to the predicate [PREDICATE] with label [LABEL].

Prompt: Write an ASP program for the following problem. Select all values associated to the predicate color with label "purple".

Encoding: select(X) :- color(X,"purple").
\end{Verbatim}
\end{scriptsize}

\spara{Filtering by negative.}
Another type of filtering criteria consists in excluding the portion of the extension of a predicate that matches with a given condition. 
This could consist, for instance, of a difference (subtraction) between two predicate extensions, but also the negation of compound conditions.   
A possible pair associating a prompt asking for filtering a predicate according to a negative condition defined over a filtered portion of a different predicate with a suitable ASP encoding is reported next.
\begin{scriptsize}
\begin{Verbatim}[frame=single, breaklines=true, breakanywhere=true, breaksymbol=$\quad$]
Template: Write an ASP program for the following problem. Select all values associated with predicate [PREDICATE_1] but not associated with predicate [PREDICATE_2] and label [LABEL].

Prompt: Write an ASP program for the following problem. Select all values associated with predicate vehicle but not associated with predicate moto and label "kawasaki".

Encoding: select(X) :- vehicle(X), 
                       not moto(X,"kawasaki").
\end{Verbatim}
\end{scriptsize}

\spara{Filtering by numeric comparisons.}
Filtering portion of tables according to the results of comparisons between terms is another typical task.
A possible pair associating a prompt asking for filtering a predicate according to a numeric comparison with a suitable ASP encoding is reported next.
\begin{scriptsize}
\begin{Verbatim}[frame=single, breaklines=true, breakanywhere=true, breaksymbol=$\quad$]
Template: Write an ASP program for the following problem. Select all values associated with predicate [PREDICATE] with a value greater or equal than [VALUE].

Prompt: Write an ASP program for the following problem. 
    Select all values associated with predicate
    size with a value greater or equal than 5.

Encoding: select(X) :- size(X,C), C>=5.
\end{Verbatim}
\end{scriptsize}

%
%
This is a small yet rich set of fundamental pieces of knowledge that an ASP user can draw upon and combine so to build more complex programs for solving different problems. 
As an example, think of many combinatorial problems, that basically consist in selecting or associating elements from a given collection in ways that comply with a given set of constraints; suitable programs for such problems are easily obtainable by combining \textbf{{\em guessing assignments}} and \textbf{{\em expressing constraints}}, possibly relying on the other types of tasks introduced above.
If also optimization plays a role, suitable solutions can be obtained by additionally \textbf{{\em expressing preferences}}.


\section{Experiments}\label{sec:exp}

In this section, we design a suite of experiments and evaluate their results. For the prompts described in the previous section, we aim at evaluating the following. 
\begin{itemize}
\item 
    \textbf{RQ1}: To what extent state-of-the-art pre-trained LLMs are actually capable of encoding textual problems into accurate ASP programs?    
\item 
    \textbf{RQ2}: Can LLMs be fine-tuned for generating ASP rules that comply to a textual description? 
\item 
    \textbf{RQ3}: To what extent can the output of the fine-tuned LLM be considered ``ready-to-use" for practical applications, in terms of syntactic and semantic correctness?
\item 
    \textbf{RQ4}: What are the limitations and potentials of this approach, e.g., how the prompt structure and/or the complexity of the problem affect the generated programs?
\end{itemize}
In order to ensure reproducibility, all the code and data used to perform the experiments are publicly available: \url{https://anonymous.4open.science/r/LLASP-D86C/}.

\subsection{Pre-trained LLMs}
Our initial goal is to assess the performance of several LLMs in generating ASP programs. 
Thus, in our experiments, we provide a structured evaluation of their performance in this domain.
%
%
We focus on freely available models that also exemplify the latest advancements in LLM-based generation: 
\begin{itemize}
\item 
    \textbf{ChatGPT 3.5}\footnote{\url{https://chatgpt.com/}}: released by OpenAI, it is a fine-tuned version from GPT-3.5, a language model trained to produce text, that has been optimized for conversation.
\item 
    \textbf{Copilot}\footnote{\url{https://learn.microsoft.com/en-us/copilot/overview}}: it is a conversational chat interface developed by Microsoft, built upon the language model GPT-4 and the text-to-image model DALL-E 3.
\item 
    \textbf{Gemini}~\cite{gemini}\footnote{\url{https://gemini.google.com/app}}: a new class of multimodal LLMs introduced by Google, showing remarkable capabilities across image, audio, video, and text understanding. 
\item 
    \textbf{Gemma}~\cite{gemma}\footnote{\url{https://huggingface.co/blog/gemma}}: released by the same group that introduced Gemini, Gemma is a lightweight family of LLMs that outperforms similarly sized models across different tasks, such as understanding, reasoning and safety. 
\item 
    \textbf{LLaMa2}~\cite{llama2}\footnote{\label{llama}\url{https://www.llama2.ai/}}: an LLM optimized for dialogue applications, developed by Meta. It shows promising results across various benchmarks, consistently outperforming open-source chat models.
\item 
    \textbf{LLaMa3}\footnote{\url{https://llama.meta.com/llama3/}}: an enhanced state-of-the-art version of LLaMa models, showing   improvements in tasks like reasoning, code generation and adherence to instructions.
\item 
    \textbf{Mistral}~\cite{mistral}\footnote{\url{https://chat.mistral.ai/chat}}: despite the reduced size of the first release (7B parameters), Mistral is a LLM that showed superior performance and efficiency across several benchmark tasks. 
    Since then, bigger and more powerful versions have been introduced. 
\end{itemize}

Table~\ref{tab:pretrained-llm-details} briefly compares the chosen LLMs in terms of model size, architectural details and training sources~\cite{llms-survey}. 
We can identify a group of large size models ($\geq70$B parameters) and another group of reduced size models ($<70$B parameters); 
Gemma 2B is the smallest model in the batch. 

\begin{table}[!ht]
    \centering
    \resizebox{\columnwidth}{!}{
    \begin{tabular}{lccc}
    \toprule
      \textbf{Model} & \textbf{No. Params.} & \textbf{Architecture} & {\textbf{Training Data}}\\
        \midrule
        ChatGPT 3.5
        & 175B & E-D & Online sources \\
        Copilot 
        & 1.5T & E-D & Github repositories \\
        Gemini
        & 1.6T & MoE & Docs, Books, Code \\
        Gemma
        & 2B-7B & D-Only & Docs, Maths, Code \\
        LLaMa2
        & 7B-13B & D-Only & Online sources \\
        LLaMa3
        & 8B-70B & D-Only + GQA & Online sources \\
        Mistral
        & 7B-141B & D-Only + GQA & Online sources \\
      \bottomrule
    \end{tabular}
    }
    \caption{Details on the evaluated pre-trained LLMs. 
    Number of parameters is reported in billions(B)/trillions(T). 
    E-D refers to the Encoder-Decoder architecture; D-Only is Decoder-Only; MoE is Mixture of Experts, and GQA is Grouped-Query Attention.}
    \label{tab:pretrained-llm-details}
\end{table}



\subsection{LLASP Fine-tuning}
The next phase of the proposed methodology involves fine-tuning a lightweight model and assessing it on the problem categories outlined in Section~\ref{sec:design}.
To achieve this, we build a training dataset by generating instances of the templates relative to such categories. 
Basically, for each problem class $\template_{\probclass^i}$, we produce variations of that template to generate a collection of prompts along with corresponding suitable ASP instances. 
Templates are instantiated by using a predefined set of predicates and labels, as shown in Section~\ref{sec:design}.
\begin{table}[!ht]
    \centering
    \resizebox{0.8\columnwidth}{!}{
    \begin{tabular}{lcc}
    \toprule
        \textbf{Problem} & \textbf{No. Tuples} & \textbf{Proportion (\%)} \\
        \midrule
        Assignment & 1,000,000 & 27\\
        Constraint & 500,000 & 13.5\\
        Combination & 100,000 & 2.7\\
        Join & 900,000 & 24.3\\
        Transitive closure & 100,000 & 2.7\\
        Preference & 400,000 & 10.8\\
        Value filtering & 100,000 & 2.7\\
        Negative filtering & 100,000 & 2.7\\
        Numeric filtering & 500,000 & 13.5\\
        \midrule
        \textbf{Total} & 3,700,000 & 100 \\
        \bottomrule
    \end{tabular}
    }
    \caption{Number of tuples and relative proportions of each problem within the dataset \dataset.}
    \label{tab:dataset_proportions}
\end{table}
The resulting dataset $\dataset$ contains approximately 4 million tuples, with varying proportions based on the problem type, as detailed in Table~\ref{tab:dataset_proportions}. 
The quantity of tuples per task depends on its syntactic complexity: 
tasks supporting greater variability in their corresponding ASP programs (such as the number of predicates in the rule, arity of the atoms, and ground instances) require more data during training. 
We further split $\dataset$ into a training and a validation set with a $80$--$20$ ratio, maintaining the proportions per problem.

We opted for Gemma 2B as our model of choice for fine-tuning. 
As the smallest among the available models, we anticipate its baseline performance to be relatively modest; this allows us to obtain tangible feedback on the effectiveness of the methodology.
We conduct Supervised Fine-Tuning (SFT) of \llm using the \texttt{SFTTrainer} component from the \texttt{trl} library\footnote{\url{https://huggingface.co/docs/trl}}, while leveraging QLora~\cite{qlora} for computational efficiency. 
The machine for training is an NVIDIA DGX server featuring a 20-Core Intel Xeon CPU, 256GB RAM  and 4X NVIDIA Tesla V100 32 GB/GPU.

\subsubsection{Evaluation Protocol.}\label{sec:validazione-esperimenti}
To assess the correctness of the generated ASP encodings, we rely on the ASP solver \textit{Clingo}~\cite{clingo} via the dedicated Python API~\footnote{\url{https://potassco.org/clingo/python-api/5.4/}}. We use Clingo for computing the answer sets of a program $P$, 
hence defining a function \clingo{P} = \as{P}, where \as{P} is the set (possibly empty) of answer sets of $P$. 
Given both the ASP program $\y \sim P_\llm(.|\x)$ generated by \llm from the prompt \x and the gold program \yreal associated with \x, we first build a set of facts $F_{\yreal}$ representing an instance of the textual problem \x. 
Given $P = \y \cup F_{\yreal}$ and $P^* = \yreal \cup F_{\yreal}$, we then perform the following checks. We invoke Clingo for computing \clingo{P}; if no parsing error occurs, we have a {\em syntactic} hit on \y. Further, we compute  \clingo{P} and \clingo{P^*}, comparing \as{P} with $\mathit{AS}(P^*)$; if they match, we have a {\em semantic} hit on \y.

\subsection{Results}
\begin{figure}
    \centering
    \includegraphics[width=\columnwidth]{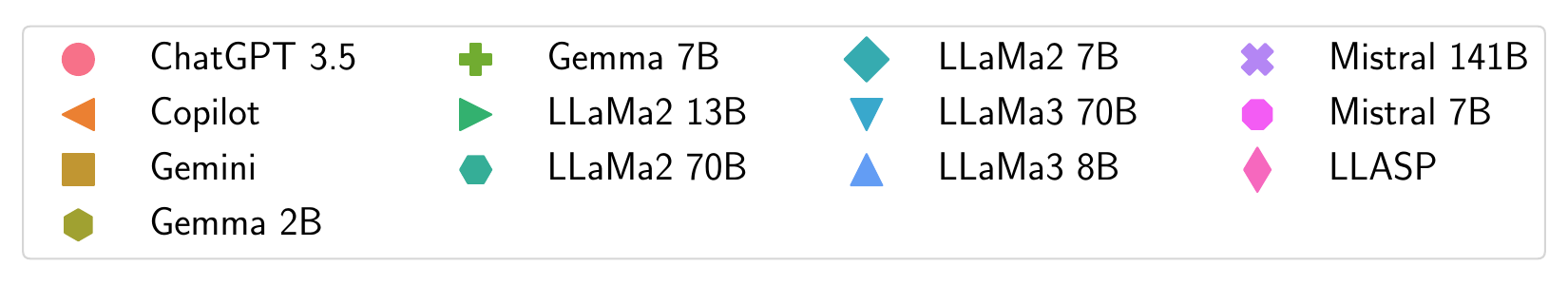}
    \includegraphics[width=0.8\columnwidth]{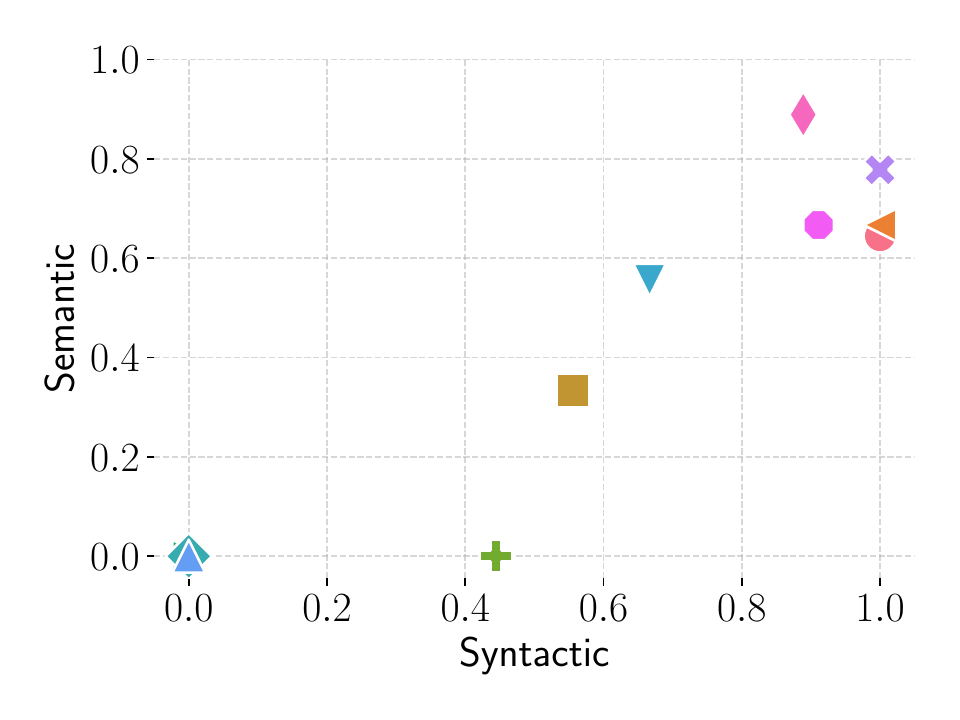}
    \caption{Comparative analysis on both syntactic and semantic results. Some markers are hidden for overlapping in (0, 0).}
    \label{fig:comparative}
\end{figure}

\begin{table}[!ht]
    \centering
    \resizebox{0.6\columnwidth}{!}{
    \begin{tabular}{lcc}
    \toprule
    \textbf{Model} & \textit{Syntactic} & \textit{Semantic} \\
    \midrule
    ChatGPT 3.5 & 1. & 0.64 \\
    Copilot & 1. & 0.67 \\
    Gemini & 0.56 & 0.33 \\
    Gemma 2B & 0. & 0. \\
    Gemma 7B & 0.44 & 0. \\
    LLaMa2 7B & 0. & 0. \\
    LLaMa2 13B & 0. & 0. \\
    LLaMa2 70B & 0. & 0. \\
    LLaMa3 8B & 0. & 0. \\
    LLaMa3 70B & 0.67 & 0.56 \\
    Mistral 7B & \underline{0.91} & 0.67 \\
    Mistral 141B & \textbf{1}. & \underline{0.78} \\
    \midrule
    \textbf{LLASP} & 0.89 & \textbf{0.89} \\
    \bottomrule
    \end{tabular}
    }
    \caption{Comparative analysis in syntactic and semantic terms.}
    \label{tab:comparative}
\end{table}





In an initial series of experiments, we conduct a comparative analysis between the pre-trained models and LLASP (i.e., the fine-tuned Gemma 2B model). 
It is noteworthy that many pre-trained models do not offer direct invocation through APIs. 
Hence, the evaluation is conducted by selecting, for each task, the example instantiation outlined in Section~\ref{sec:design}. 
Then, the corresponding prompt is submitted, and results are assessed according to the aforementioned protocol. 
The process is repeated $5$ times to ensure statistical robustness over the sampling process intrinsic of the LLMs. 

Table~\ref{tab:comparative} reports accuracy of all the considered models, considering both syntactic and semantic perspectives;  
Figure~\ref{fig:comparative} offers a concise visual representation of these findings. 
Notably, a syntactic hit does not necessarily imply with semantic correctness. For instance, Gemma 7B achieves 44\% syntactic accuracy but shows no semantic quality at all. 
In general, no model achieves complete correctness across the board, and lightweight models report the weakest performance (with the only notable exception of Mistral 7B).

Among the models, ChatGPT, Copilot, and Mistral 141B achieve 100\% syntactic accuracy. 
Surprisingly, both Gemini and LLaMa3 70B display lower accuracy with respect to these methods, despite being comparable in model size. LLASP emerges as the model with the highest semantic accuracy among all the models considered. 
We note that whenever a generation produces a syntactically correct program, it also tends to be semantically correct. 
Table~\ref{tab:comparative} confirms this observation, indicating that this phenomenon occurs only with LLASP. 
This is particularly noteworthy, given that the original Gemma 2B yields entirely inconsistent results.
\begin{table*}[!ht]
    \centering
    \resizebox{\linewidth}{!}{
    \begin{tabular}{l cc cc cc cc cc cc cc cc cc}
    \toprule
         \multicolumn{1}{c}{\textbf{Model}} & \multicolumn{2}{c}{\textbf{Assignment}} & \multicolumn{2}{c}{\textbf{Constraint}} & \multicolumn{2}{c}{\textbf{Combination}} & \multicolumn{2}{c}{\textbf{Join}} & \multicolumn{2}{c}{\textbf{Closure}} & \multicolumn{2}{c}{\textbf{Preference}} & \multicolumn{2}{c}{\textbf{Value Filter}} & \multicolumn{2}{c}{\textbf{Neg. Filter}} & \multicolumn{2}{c}{\textbf{Num. Filter}} \\
         \cmidrule(lr){2-3}
         \cmidrule(lr){4-5}
         \cmidrule(lr){6-7}
         \cmidrule(lr){8-9}
         \cmidrule(lr){10-11}
         \cmidrule(lr){12-13}
         \cmidrule(lr){14-15}
         \cmidrule(lr){16-17}
         \cmidrule(lr){18-19}
         & \textit{Syn.} & \textit{Sem.} & \textit{Syn.} & \textit{Sem.} & \textit{Syn.} & \textit{Sem.} & \textit{Syn.} & \textit{Sem.} & \textit{Syn.} & \textit{Sem.} & \textit{Syn.} & \textit{Sem.} & \textit{Syn.} & \textit{Sem.} & \textit{Syn.} & \textit{Sem.} & \textit{Syn.} & \textit{Sem.} \\ 
         \midrule
         ChatGPT 3.5 & 1. & 0.8 & 1. & 1. & 1. & 1. & 1.  & 1. & 1. & 1. & 1. & 0. & 1. & 0. & 1. & 0. & 1. & 1. \\
         Copilot & 1. & 0. & 1. & 1. & 1. & 1. & 1. & 1. & 1. & 1. & 1. & 0. & 1. & 1. & 1. & 0. & 1. & 1. \\
         Gemini & 0. & 0. & 1. & 0. & 0. & 0. & 1. & 1. & 1. & 1. & 0. & 0. & 0.6 & 0. & 0.4 & 0. & 1. & 1. \\
         Gemma 2B & 0. & 0. & 0. & 0. & 0. & 0. & 0. & 0. & 0. & 0. & 0. & 0. & 0. & 0. & 0. & 0. & 0. & 0. \\
         Gemma 7B & 0. & 0. & 1. & 0. & 0. & 0. & 1. & 0. & 0. & 0. & 0. & 0. & 0. & 0. & 1. & 0. & 1. & 0.\\
         LLaMa2 7B & 0. & 0. & 0. & 0. & 0. & 0. & 0. & 0. & 0. & 0. & 0. & 0. & 0. & 0. & 0. & 0. & 0. & 0. \\
         LLaMa2 13B & 0. & 0. & 0. & 0. & 0. & 0. & 0. & 0. & 0. & 0. & 0. & 0. & 0. & 0. & 0. & 0. & 0. & 0. \\
         LLaMa2 70B & 0. & 0. & 0. & 0. & 0. & 0. & 0. & 0. & 0. & 0. & 0. & 0. & 0. & 0. & 0. & 0. & 0. & 0. \\
         LLaMa3 8B & 0. & 0. & 0. & 0. & 0. & 0. & 0. & 0. & 0. & 0. & 0. & 0. & 0. & 0. & 0. & 0. & 0. & 0. \\
         LLaMa3 70B & 0.8 & 0.2 & 0.4 & 0.4 & 1. & 1. & 0.6 & 0.6 & 1. & 1. & 0. & 0. & 0.6 & 0.6 & 0.6 & 0.2 & 1. & 1. \\
         Mistral 7B & 0.8 & 0. & 1. & 0. & 1. & 1. & 1. & 1. & 1. & 1. & 0.4 & 0. & 1. & 1. &  1. & 1. & 1. & 1. \\
         Mistral 141B & 1. & 0. & 1. & 1. & 1. & 1. & 1. & 1. & 1. & 1. & 1. & 1. & 1. & 1. & 1. & 0. & 1. & 1. \\
         \midrule
         \textbf{LLASP} & 1. & 1. & 1. & 1. & 1. & 1. & 0. & 0. & 1. & 1. & 1. & 1. & 1. & 1. & 1. & 1. & 1. & 1.\\
         \bottomrule
    \end{tabular}
    }
    \caption{Generation accuracy in terms of syntactic (\textit{Syn.}) and semantic (\textit{Sem.}) perspective across all the considered LLMs.}
    \label{tab:syntactic-semantic-scores}
\end{table*}
Table~\ref{tab:syntactic-semantic-scores} further details the results for each task. 
Compared to large sized models, LLASP only fails on the join task, due to syntactic fails. 
By contrast, Mistral 141B exhibits semantic failures both on assignment and negative filtering. 
\begin{table}[!ht]
    \centering
    \resizebox{0.6\columnwidth}{!}{
    \begin{tabular}{lcc}
    \toprule
         \textbf{Model} & \textit{Syntactic} & \textit{Semantic} \\
         \midrule
         ChatGPT 3.5 & \textbf{1.} & 0.67 \\
         Copilot & \textbf{1.} & 0.57 \\
         Mistral 7B & \textbf{1.} & \underline{0.85} \\
         Mistral 141B & \textbf{1.} & 0.69 \\
         \midrule
         \textbf{LLASP} & \underline{0.93} & \textbf{0.93} \\
         \bottomrule
    \end{tabular}
    }
    \caption{In-depth comparison using an extended dataset. }
    \label{tab:comparison-with-best}
\end{table}

We provide more insights regarding the generation flaws via anecdotal evidence of the ASP encoding provided by the identified top-performing models (see Figure~\ref{fig:comparative}): Mistral 141B, ChatGPT 3.5, Copilot, and Mistral 7B. 
In particular, we focus on semantic correctness of the generated programs, which in our context is the most representative quality measure. 
Surprisingly, almost all of the aforesaid model, except for ChatGPT 3.5, fail in representing the guessing assignment via a semantically correct ASP encoding, producing an average score of 0 out of 5 (see Table~\ref{tab:syntactic-semantic-scores}). 
ChatGPT 3.5, indeed generated a semantically correct program 4 times out of 5, achieving a score of $0.8$.
We report below the encoding produced by each of the models, comparing it with the one generated by LLASP, in the response to the Guessing Assignment task reported in Section~\ref{sec:design}.

\spara{Mistral 7B/141B.} Despite being syntactically correct, the generated program presents a semantic flow, since it does not allow multiple assignments for the same label. This property is not requested in the prompt, and would lead to inconsinstency whenever the number of labels is lower than the number of predicates. 
\begin{scriptsize}
    \begin{Verbatim}[frame=single, breaklines=true, breakanywhere=true, breaksymbol=$\quad$, commandchars=\\\{\}]
Encoding: 
    1 \{ assigned(X, L) : label(L) \} 1 :- city(X).    
    :- assigned(X1, L), assigned(X2, L), X1 != X2.
    \end{Verbatim}
\end{scriptsize}

\spara{ChatGPT 3.5.} Similarly, the result of the generation is a syntactically correct ASP encoding, according to the Clingo solver. However, it fails in capturing the semantics of the textual prompt. 

\begin{scriptsize}
    \begin{Verbatim}[frame=single, breaklines=true, breakanywhere=true, breaksymbol=$\quad$, commandchars=\\\{\}]
Encoding: 
    1 \{ assign(X, L) : label(L) \} 1 :- city(X).
    :- assign(X, L1), assign(Y, L1), X != Y, label(L1).
    :- city(X), not assign(X, _).
    \end{Verbatim}
\end{scriptsize}
    
\spara{Copilot.} Here, the generated program encodes a completely different semantics with respect to the problem description: indeed, the above fragment specifies a join between the predicate ``city" and the predicate ``label" via the ``City" label. 
\begin{scriptsize}
    \begin{Verbatim}[frame=single, breaklines=true, breakanywhere=true, breaksymbol=$\quad$, commandchars=\\\{\}]
Encoding: 
    assign_label(City, Label) :-
        city(City), label(City, Label).
    \end{Verbatim}
\end{scriptsize}

\spara{LLASP.} Eventually, we report an example of wrong ASP program produced by LLASP over the join task discussed in Section~\ref{sec:design}. 
The generated encoding presents a syntactic error  regarding both the arity of the predicate ``owner" and  the presence of the ``Z" predicate, thus affecting the semantics of the program.
\begin{scriptsize}
    \begin{Verbatim}[frame=single, breaklines=true, breakanywhere=true, breaksymbol=$\quad$, commandchars=\\\{\}]    
Encoding: owner_restaurant(X,Z):-owner(X,Y),Z(Y).
    \end{Verbatim}
\end{scriptsize}

As an additional test, in order to further assess the robustness of the findings, we compare LLASP against the aforesaid models by performing a more detailed evaluation on an extended set of test instances, where prompts vary in terms of predicates and labels. 
Table~\ref{tab:comparison-with-best} reports the results. 
Again, despite occasional syntactic flaws, LLASP offers more reliable encodings, outperforming the second-best model (Mistral 7B) of almost ten percentage points. 
Moreover, we observe that the problem semantics is fulfilled whenever syntactic correctness is ensured. 

Encouraged by these results, we proceeded in testing LLASP on a larger scale. 
To this purpose, we generated a test set $\dataset_{test}$ consisting of 9,000 prompts, following the same principles as the training set $\dataset$ described earlier, yet, again, on a different set of predicates and labels. 
In this test set, the number of examples for each task is evenly distributed. 
Table~\ref{tab:pretrained_llm} presents the accuracy of LLASP over $\dataset_{test}$. 
We notice that: (\textit{i}) unlike in the previous experiment, the consistency between syntactic and semantic accuracy is not upheld; (\textit{ii}) the quality of the generated programs remains consistently high over all the tasks, even on the join problem which failed in our first experiment. 
Given the substantial scale of $\dataset_{test}$, we are confident that this experiment demonstrates the resilience of the proposed approach.
\begin{table}[!ht]
    \centering
    \resizebox{0.7\columnwidth}{!}{
    \begin{tabular}{lcc}
    \toprule
      \textbf{Problem} &  \textit{Syntactic} &  \textit{Semantic} \\
        \midrule
         Assignment & 0.76 & 0.73 \\
         Constraint & 1. & 1. \\
         Combination & 1. & 0.81 \\
         Join & 0.95 & 0.91 \\
         Closure & 1. & 0.99 \\
         Preference & 1. & 0.88 \\
         Value Filtering & 1. & 0.89 \\
         Negative Filtering & 1. & 0.89 \\
         Numeric Filtering & 1. & 0.9 \\
         \midrule
         \textbf{Total} (avg) & \textbf{0.97} & \textbf{0.89} \\ 
      \bottomrule
    \end{tabular}
    }
    \caption{Results in terms of syntactic and semantic generation accuracy of LLASP over $\dataset_{test}$.}
    \label{tab:pretrained_llm}
\end{table}

\subsection{Limitations}
\label{sec:limitations}
To explore the ability of generating complex ASP programs from atomic tasks, we conducted further experiments for highlighting limitations and potentials, by fine-tuning LLASP on an expanded version of the dataset \dataset, incorporating combined prompts. 
The results are mixed. 
While we observe a surprisingly successful generation rate for several complex combinations, we also witness failures on other simple ones. 
We believe that this is due to a suboptimal design of the training set regarding such combinations; 
indeed, we employed random combinations, whereas a more strategic approach should be devised. 

We further assessed the generation's generalization capabilities through a simple preliminary experiment:
we generated and included $10$ possible rephrasings for a given prompt in the dataset. 
However, the generation results were unsuccessful, suggesting that a more comprehensive evaluation requires a careful design of the training dataset.


Additionally, we explored various training strategies, including Reinforcement Learning with Human Feedback (RLHF)~\cite{rlhf}. 
In this setup, generation is linked to a reward that considers both syntactic and semantic aspects of the ASP program generated. 
However, the results did not surpass those of standard SFT training. 
We posit that a more meticulous design of the reward function is needed, were several aspects of the generation (e.g., rule redundancy, fluency or verbosity) are be considered. 


\section{Conclusions and Future Work}\label{sec:conc}
In this study, we conducted an extensive examination of the capability of Large Language Models (LLMs) to accurately encode Answer Set Programming (ASP) programs based on natural language specifications. 
To the best of our knowledge, this is the first extensive comparison of state-of-the-art LLMs in terms of ASP code generation. 
Our findings show that, despite the large potential, LLMs are not ready for such a task, as they underperform in terms of programs correctness (both syntactic and semantic). 
With this respect, we demonstrate that tailored training can significantly mitigate this deficiency, offering a pathway towards fine-tuned models able to generate ASP programs ready for practical deployment. 
Indeed, we propose LLASP, a fine-tuned version of the lightweight Gemma 2B base-model. We train it over a comprehensive datased we designed and populated on purpose so to capture ASP fundamental patterns. LLASP-generated programs proved to be significantly better than those generated by larger and more powerful LLMs, especially under a semantic perspective. 

We also devised and performed a significant set of additional experiments, that provide interesting insights on the topic and suggest that this study can be expanded in multiple directions. In particular, besides the pointers already discussed in Section~\ref{sec:limitations} (i.e., generalizability, complex prompts, RLHF training), we claim that a novel metric is necessary, to be used both in training and evaluation. 
In principle, this metric should take into account both 
syntactic correctness and capability  of the generated ASP programs to effectively tackle problems described by the prompts in the first place. 
%
%
Furthermore, we aim at developing novel controlled training methodologies to incorporate ASP syntactic guidance within the training process of the LLMs. 
This would also require a careful inspection of the underlying architecture and the adoption of suitable modifications. 


\begin{thebibliography}{}

\bibitem[\protect\citeauthoryear{Baral and Dzifcak}{2011}]{DBLP:conf/aaaifs/BaralD11}
Baral, C., and Dzifcak, J.
\newblock 2011.
\newblock Solving puzzles described in english by automated translation to answer set programming and learning how to do that translation.
\newblock In {\em {AAAI} Fall Symposium: Advances in Cognitive Systems}, volume {FS-11-01} of {\em {AAAI} Technical Report}.
\newblock {AAAI}.

\bibitem[\protect\citeauthoryear{Baral}{2010}]{DBLP:books/daglib/0040913}
Baral, C.
\newblock 2010.
\newblock {\em Knowledge Representation, Reasoning and Declarative Problem Solving}.
\newblock Cambridge University Press.

\bibitem[\protect\citeauthoryear{Borroto, Kareem, and Ricca}{2024}]{ijcai24-borroto}
Borroto, M.; Kareem, I.; and Ricca, F.
\newblock 2024.
\newblock Towards automatic composition of {ASP} programs from natural language specifications.
\newblock In {\em {IJCAI}}, volume abs/2403.04541,  to appear.
\newblock ijcai.org.

\bibitem[\protect\citeauthoryear{Brewka, Eiter, and Truszczynski}{2011}]{DBLP:journals/cacm/BrewkaET11}
Brewka, G.; Eiter, T.; and Truszczynski, M.
\newblock 2011.
\newblock Answer set programming at a glance.
\newblock {\em Commun. {ACM}} 54(12):92--103.

\bibitem[\protect\citeauthoryear{Calimeri \bgroup et al\mbox.\egroup }{2016}]{DBLP:journals/ai/CalimeriGMR16}
Calimeri, F.; Gebser, M.; Maratea, M.; and Ricca, F.
\newblock 2016.
\newblock Design and results of the fifth answer set programming competition.
\newblock {\em Artif. Intell.} 231:151--181.

\bibitem[\protect\citeauthoryear{Caruso \bgroup et al\mbox.\egroup }{2024}]{DBLP:journals/tplp/CarusoDMMR24}
Caruso, S.; Dodaro, C.; Maratea, M.; Mochi, M.; and Riccio, F.
\newblock 2024.
\newblock {CNL2ASP:} converting controlled natural language sentences into {ASP}.
\newblock {\em Theory Pract. Log. Program.} 24(2):196--226.

\bibitem[\protect\citeauthoryear{Chen, Tworek, and et al.}{2021}]{DBLP:journals/corr/abs-2107-03374}
Chen, M.; Tworek, J.; and et~al., H.~J.
\newblock 2021.
\newblock Evaluating large language models trained on code.
\newblock {\em CoRR} abs/2107.03374.

\bibitem[\protect\citeauthoryear{Dakhel \bgroup et al\mbox.\egroup }{2023}]{DBLP:journals/jss/DakhelMNKDJ23}
Dakhel, A.~M.; Majdinasab, V.; Nikanjam, A.; Khomh, F.; Desmarais, M.~C.; and Jiang, Z. M.~J.
\newblock 2023.
\newblock Github copilot {AI} pair programmer: Asset or liability?
\newblock {\em J. Syst. Softw.} 203:111734.

\bibitem[\protect\citeauthoryear{Dettmers \bgroup et al\mbox.\egroup }{2023}]{qlora}
Dettmers, T.; Pagnoni, A.; Holtzman, A.; and Zettlemoyer, L.
\newblock 2023.
\newblock Qlora: Efficient finetuning of quantized llms.

\bibitem[\protect\citeauthoryear{Eiter, Ianni, and Krennwallner}{2009}]{DBLP:conf/rweb/EiterIK09}
Eiter, T.; Ianni, G.; and Krennwallner, T.
\newblock 2009.
\newblock Answer set programming: {A} primer.
\newblock In Tessaris, S.; Franconi, E.; Eiter, T.; Gutierrez, C.; Handschuh, S.; Rousset, M.; and Schmidt, R.~A., eds., {\em Reasoning Web. Semantic Technologies for Information Systems, 5th International Summer School 2009, Brixen-Bressanone, Italy, August 30 - September 4, 2009, Tutorial Lectures}, volume 5689 of {\em Lecture Notes in Computer Science},  40--110.
\newblock Springer.

\bibitem[\protect\citeauthoryear{Erdem and Yeniterzi}{2009}]{DBLP:conf/bionlp/ErdemY09}
Erdem, E., and Yeniterzi, R.
\newblock 2009.
\newblock Transforming controlled natural language biomedical queries into answer set programs.
\newblock In {\em BioNLP@HLT-NAACL},  117--124.
\newblock ACL.

\bibitem[\protect\citeauthoryear{Ernst and Bavota}{2022}]{DBLP:journals/software/ErnstBM22}
Ernst, N.~A., and Bavota, G.
\newblock 2022.
\newblock Ai-driven development is here: Should you worry?
\newblock {\em {IEEE} Softw.} 39(2):106--110.

\bibitem[\protect\citeauthoryear{Fang and Tompits}{2017}]{DBLP:conf/inap/FangT17}
Fang, M., and Tompits, H.
\newblock 2017.
\newblock An approach for representing answer sets in natural language.
\newblock In {\em {DECLARE}}, volume 10997 of {\em LNCS},  115--131.
\newblock Springer.

\bibitem[\protect\citeauthoryear{Gebser \bgroup et al\mbox.\egroup }{2012}]{DBLP:series/synthesis/2012Gebser}
Gebser, M.; Kaminski, R.; Kaufmann, B.; and Schaub, T.
\newblock 2012.
\newblock {\em Answer Set Solving in Practice}.
\newblock Synthesis Lectures on Artificial Intelligence and Machine Learning. Morgan {\&} Claypool Publishers.

\bibitem[\protect\citeauthoryear{Gebser \bgroup et al\mbox.\egroup }{2016}]{clingo}
Gebser, M.; Kaminski, R.; Kaufmann, B.; Ostrowski, M.; Schaub, T.; and Wanko, P.
\newblock 2016.
\newblock Theory solving made easy with clingo 5.
\newblock In {\em International Conference on Logic Programming}.

\bibitem[\protect\citeauthoryear{Gebser \bgroup et al\mbox.\egroup }{2018}]{DBLP:conf/ijcai/GebserLMPRS18}
Gebser, M.; Leone, N.; Maratea, M.; Perri, S.; Ricca, F.; and Schaub, T.
\newblock 2018.
\newblock Evaluation techniques and systems for answer set programming: a survey.
\newblock In Lang, J., ed., {\em Proceedings of the Twenty-Seventh International Joint Conference on Artificial Intelligence, {IJCAI} 2018, July 13-19, 2018, Stockholm, Sweden},  5450--5456.
\newblock ijcai.org.

\bibitem[\protect\citeauthoryear{Gelfond and Lifschitz}{1991}]{DBLP:journals/ngc/GelfondL91}
Gelfond, M., and Lifschitz, V.
\newblock 1991.
\newblock Classical negation in logic programs and disjunctive databases.
\newblock {\em New Gener. Comput.} 9(3/4):365--386.

\bibitem[\protect\citeauthoryear{Ishay, Yang, and Lee}{2023}]{DBLP:conf/kr/IshayY023}
Ishay, A.; Yang, Z.; and Lee, J.
\newblock 2023.
\newblock Leveraging large language models to generate answer set programs.
\newblock In {\em {KR}},  374--383.

\bibitem[\protect\citeauthoryear{Jiang \bgroup et al\mbox.\egroup }{2023}]{mistral}
Jiang, A.~Q.; Sablayrolles, A.; Mensch, A.; Bamford, C.; Chaplot, D.~S.; de~las Casas, D.; Bressand, F.; Lengyel, G.; Lample, G.; Saulnier, L.; Lavaud, L.~R.; Lachaux, M.-A.; Stock, P.; Scao, T.~L.; Lavril, T.; Wang, T.; Lacroix, T.; and Sayed, W.~E.
\newblock 2023.
\newblock Mistral 7b.

\bibitem[\protect\citeauthoryear{Kalliamvakou}{2022}]{kalliamvakou2022research}
Kalliamvakou, E.
\newblock 2022.
\newblock Research: quantifying github copilot’s impact on developer productivity and happiness.
\newblock {\em The GitHub Blog}.

\bibitem[\protect\citeauthoryear{Kuhn}{2014}]{DBLP:journals/coling/Kuhn14}
Kuhn, T.
\newblock 2014.
\newblock A survey and classification of controlled natural languages.
\newblock {\em Comput. Linguistics} 40(1):121--170.

\bibitem[\protect\citeauthoryear{Lifschitz}{2019}]{Lifschitz19}
Lifschitz, V.
\newblock 2019.
\newblock {\em Answer Set Programming}.
\newblock Springer.

\bibitem[\protect\citeauthoryear{Ma \bgroup et al\mbox.\egroup }{2024}]{llamoco}
Ma, Z.; Guo, H.; Chen, J.; Peng, G.; Cao, Z.; Ma, Y.; and Gong, Y.-J.
\newblock 2024.
\newblock Llamoco: Instruction tuning of large language models for optimization code generation.

\bibitem[\protect\citeauthoryear{Minaee \bgroup et al\mbox.\egroup }{2024}]{llms-survey}
Minaee, S.; Mikolov, T.; Nikzad, N.; Chenaghlu, M.; Socher, R.; Amatriain, X.; and Gao, J.
\newblock 2024.
\newblock Large language models: A survey.

\bibitem[\protect\citeauthoryear{Mitra and Baral}{2016}]{DBLP:conf/aaai/MitraB16}
Mitra, A., and Baral, C.
\newblock 2016.
\newblock Addressing a question answering challenge by combining statistical methods with inductive rule learning and reasoning.
\newblock In {\em {AAAI}},  2779--2785.
\newblock {AAAI} Press.

\bibitem[\protect\citeauthoryear{Naveed \bgroup et al\mbox.\egroup }{2024}]{comprehensive-overview}
Naveed, H.; Khan, A.~U.; Qiu, S.; Saqib, M.; Anwar, S.; Usman, M.; Akhtar, N.; Barnes, N.; and Mian, A.
\newblock 2024.
\newblock A comprehensive overview of large language models.

\bibitem[\protect\citeauthoryear{Nye \bgroup et al\mbox.\egroup }{2021}]{DBLP:conf/nips/NyeTTL21}
Nye, M.~I.; Tessler, M.~H.; Tenenbaum, J.~B.; and Lake, B.~M.
\newblock 2021.
\newblock Improving coherence and consistency in neural sequence models with dual-system, neuro-symbolic reasoning.
\newblock In {\em NeurIPS},  25192--25204.

\bibitem[\protect\citeauthoryear{Peng \bgroup et al\mbox.\egroup }{2023}]{DBLP:journals/corr/abs-2302-06590}
Peng, S.; Kalliamvakou, E.; Cihon, P.; and Demirer, M.
\newblock 2023.
\newblock The impact of {AI} on developer productivity: Evidence from github copilot.
\newblock {\em CoRR} abs/2302.06590.

\bibitem[\protect\citeauthoryear{Raiaan \bgroup et al\mbox.\egroup }{2024}]{survey}
Raiaan, M. A.~K.; Mukta, M. S.~H.; Fatema, K.; Fahad, N.~M.; Sakib, S.; Mim, M. M.~J.; Ahmad, J.; Ali, M.~E.; and Azam, S.
\newblock 2024.
\newblock A review on large language models: Architectures, applications, taxonomies, open issues and challenges.
\newblock {\em IEEE Access} 12:26839--26874.

\bibitem[\protect\citeauthoryear{Rajasekharan \bgroup et al\mbox.\egroup }{2023}]{reliable-nlu}
Rajasekharan, A.; Zeng, Y.; Padalkar, P.; and Gupta, G.
\newblock 2023.
\newblock Reliable natural language understanding with large language models and answer set programming.
\newblock {\em Electronic Proceedings in Theoretical Computer Science} 385:274–287.

\bibitem[\protect\citeauthoryear{Schwitter}{2018}]{DBLP:journals/tplp/Schwitter18}
Schwitter, R.
\newblock 2018.
\newblock Specifying and verbalising answer set programs in controlled natural language.
\newblock {\em Theory Pract. Log. Program.} 18(3-4):691--705.

\bibitem[\protect\citeauthoryear{Stahlberg}{2020}]{DBLP:journals/jair/Stahlberg20}
Stahlberg, F.
\newblock 2020.
\newblock Neural machine translation: {A} review.
\newblock {\em J. Artif. Intell. Res.} 69:343--418.

\bibitem[\protect\citeauthoryear{Team}{2024a}]{gemini}
Team, G.
\newblock 2024a.
\newblock Gemini: A family of highly capable multimodal models.

\bibitem[\protect\citeauthoryear{Team}{2024b}]{gemma}
Team, G.
\newblock 2024b.
\newblock Gemma: Open models based on gemini research and technology.

\bibitem[\protect\citeauthoryear{Touvron and others}{2023}]{llama2}
Touvron, H., et~al.
\newblock 2023.
\newblock Llama 2: Open foundation and fine-tuned chat models.

\bibitem[\protect\citeauthoryear{Vaswani \bgroup et al\mbox.\egroup }{2017}]{attention-is-all}
Vaswani, A.; Shazeer, N.; Parmar, N.; Uszkoreit, J.; Jones, L.; Gomez, A.~N.; Kaiser, L.~u.; and Polosukhin, I.
\newblock 2017.
\newblock Attention is all you need.
\newblock In Guyon, I.; Luxburg, U.~V.; Bengio, S.; Wallach, H.; Fergus, R.; Vishwanathan, S.; and Garnett, R., eds., {\em Advances in Neural Information Processing Systems}, volume~30.
\newblock Curran Associates, Inc.

\bibitem[\protect\citeauthoryear{Wang \bgroup et al\mbox.\egroup }{2023}]{codet5}
Wang, Y.; Le, H.; Gotmare, A.~D.; Bui, N. D.~Q.; Li, J.; and Hoi, S. C.~H.
\newblock 2023.
\newblock Codet5+: Open code large language models for code understanding and generation.

\bibitem[\protect\citeauthoryear{Xu \bgroup et al\mbox.\egroup }{2022}]{evaluation-code}
Xu, F.~F.; Alon, U.; Neubig, G.; and Hellendoorn, V.~J.
\newblock 2022.
\newblock A systematic evaluation of large language models of code.
\newblock In {\em Proceedings of the 6th ACM SIGPLAN International Symposium on Machine Programming}, MAPS 2022,  1–10.
\newblock New York, NY, USA: Association for Computing Machinery.

\bibitem[\protect\citeauthoryear{Yang, Ishay, and Lee}{2023}]{DBLP:conf/acl/YangI023}
Yang, Z.; Ishay, A.; and Lee, J.
\newblock 2023.
\newblock Coupling large language models with logic programming for robust and general reasoning from text.
\newblock In {\em {ACL} (Findings)},  5186--5219.
\newblock ACL.

\bibitem[\protect\citeauthoryear{Ziegler \bgroup et al\mbox.\egroup }{2020}]{rlhf}
Ziegler, D.~M.; Stiennon, N.; Wu, J.; Brown, T.~B.; Radford, A.; Amodei, D.; Christiano, P.; and Irving, G.
\newblock 2020.
\newblock Fine-tuning language models from human preferences.

\end{thebibliography}

\end{document}